\documentclass{article}

\PassOptionsToPackage{square,numbers}{natbib}

\usepackage[final]{neurips_2025_ml4ps}

\usepackage[utf8]{inputenc} 
\usepackage[T1]{fontenc}    
\usepackage{hyperref}       
\usepackage{url}            
\usepackage{booktabs}       
\usepackage{amsfonts}       
\usepackage{nicefrac}       
\usepackage{microtype}      
\usepackage{xcolor}         

\usepackage{amsmath}        
\usepackage{amsfonts}
\usepackage{dsfont}
\usepackage{lipsum} 
\usepackage{graphicx}
\usepackage{subcaption}
\usepackage{wrapfig}
\usepackage{nameref}
\usepackage{algorithm}
\usepackage{algorithmic}
\usepackage{minitoc}

\graphicspath{{figs/}}

\title{Power Ensemble Aggregation for\\ Improved Extreme Event AI Prediction}

\author{
  Julien Collard\\
  Ecole Polytechnique\\
  Rte de Saclay, 91120 Palaiseau, France\\
  \texttt{julien.collard@polytechnique.org}\\
  \and
  Pierre Gentine\\
  LEAP, Columbia University\\
  2276 12th ave, New York, NY 10027, USA\\
  \texttt{pg2328@columbia.edu}\\
  \and
  Tian Zheng\\
  Department of Statistics, Columbia University\\
  1255 Amsterdam Avenue, New York, NY 10027, USA\\
}

\begin{document}

\maketitle

\begin{abstract}
  This paper addresses the critical challenge of improving predictions of climate extreme events, specifically heat waves, using machine learning methods. 
  Our work is framed as a classification problem in which we try to predict whether surface air temperature will exceed its q-th local quantile within a specified timeframe. 
  Our key finding is that aggregating ensemble predictions using a power mean significantly enhances the classifier's performance. 
  By making a machine-learning based weather forecasting model generative and applying this non-linear aggregation method, we achieve better accuracy in predicting extreme heat events than with the typical mean prediction from the same model. 
  Our power aggregation method shows promise and adaptability, as its optimal performance varies with the quantile threshold chosen, demonstrating increased effectiveness for higher extremes prediction.
\end{abstract}

\section{Introduction}

Numerical weather prediction (NWP) has recently made significant progress with the development of deep learning models \citep{biPanguWeather3DHighResolution2022,pathakFourCastNetGlobalDatadriven2022, lamGraphCastLearningSkillful2023}.
With the emergence of new architectures and the use of massive datasets \citep{hersbachERA5GlobalReanalysis2020} and well-defined benchmarks \citep{raspWeatherBenchBenchmarkNext2024}, these models have been able to predict weather variables more and more accurately and now usually outperform physically based models \citep{raspWeatherBenchBenchmarkNext2024}.
However, predicting extreme events remains a challenge, as they are rare by definition and follow different patterns than the rest of the data. \citep{zscheischlerTypologyCompoundWeather2020,davisonStatisticsExtremes2015}
For those events machine-learning based models are still struggling as recently shown by \citet{zhang2025numericalmodelsoutperformai} and \citet{Sun_2025}.
Therefore, predicting extremes with ML requires specialized techniques and is an active field of research \citep{fangSurveyApplicationDeep2021, magnussonVerificationExtremeWeather201409, lopez-gomezGlobalExtremeHeat2023}.

In this study, we investigate an adaptive ensemble aggregation method to predict extreme events applicable to any generative model \citep{chenGenerativeMachineLearning2024,berrischCRPSLearning2023}.
This method, based on the power mean \citep{cantrellPowerMean,borweinPiAGMStudy1998} was previously tested in different fields \citep{hassanNewMethodEnsemble2021}.
We applied it to extreme surface air temperature classification, but this method could be adapted to any extreme prediction problem.
For our experiments, we used a custom NWP generative model inspired by \citet{weynImprovingDataDrivenGlobal2020} and \citet{lopez-gomezGlobalExtremeHeat2023} based on convolutions on a cubic sphere grid and made generative using Perlin noise \citep{perlinImageSynthesizer1985a} for input perturbation.

\section{Methodology}

\subsection{Data}

For this study, we used the ERA5 reanalysis data from the European Centre for Medium-Range Weather Forecasts (ECMWF) \citep{hersbachERA5GlobalReanalysis2020} for training and evaluation. 
We downloaded data resampled at 1.5° spatial resolution, covering 1990-2020 and at 6-hour temporal resolution using the WeatherBench2 \citep{raspWeatherBenchBenchmarkNext2024} cloud-based datasets.
We chose to resample the data to a daily temporal resolution to remove the diurnal cycle and focus on large temporal scale extremes.

The full list of variables downloaded for this study and the levels at which they were sampled are presented in appendix in Table \ref{tab:variables}. This choice of weather variables to predict surface air temperature was inspired by the work of \citet{lopez-gomezGlobalExtremeHeat2023}. 

Following the work of \citet{weynImprovingDataDrivenGlobal2020}, we regridded the data to a 6x48x48 gnomonic equiangular cube sphere grid (appendix \ref{app:cube_sphere_grid}) to avoid pole singularities and allow the use of the CubeSphere Convolutional Neural Networks they introduced.

\subsection{Defining extreme events}

To define extremes, we used local climatologies, i.e. weather variable distributions defined for each location and time of the year. 

For a given threshold $ 0.5 \leq q \leq 1 $, we define an extreme heat event as the exceedance of the $q$-th local quantile of the surface air temperature.
This local definition is essential as we want to capture extreme events happening at any location and time of the year.
Without it, we would be limited to capturing global extremes in hot regions like the Sahara desert.

Assuming that surface air temperatures nearly follow a Gaussian distribution, a temperature is considered $q$-extreme if its local anomaly $x$ is such that $\Phi(x) \geq q$ with $\Phi$ the cumulative distribution function of the standard normal distribution $\mathcal{N}(0,1)$. Local anomalies are computed by subtracting local climatology mean to temperatures and dividing by local climatology standard deviation. 

\subsection{Classification problem}

In this study, we consider extreme events' detection as a classification problem:
knowing the state of the atmosphere on a given day $d$ and its past few days, we want to predict whether the surface air temperature at any location will exceed its $q$-th local quantile on day $d+\Delta d$.
For a given quantile $q$ to be exceeded, on a given day $d$, a given time lead $\Delta d$ and a given location, this becomes a binary classification problem.
We denote $x$ as the true local anomaly of the surface air temperature at time $d+\Delta d$ and at this location.
The ground truth label $y$ is then defined as $y = \mathds{1}_{\Phi(x) \geq q} \in \{0,1\}$.

We expect models to predict a score $\hat{s} \in [0,1]$ that reflects how confident they are that the surface air temperature will exceed its $q$-th local quantile.
Such a model is called a "score-based classifier".
For real applications, a threshold $\tau$ must then be chosen to convert the score into a binary prediction $\hat{y} = \mathds{1}_{\hat{s} \geq \tau}$.

To evaluate the performance of the heat-wave classifiers, we used the area under the receiver operating characteristic curve (AUC), a well-established metric for binary classification problems.
It is defined as the area under the curve of the true positive rate (TPR) as a function of the false positive rate (FPR) for different thresholds.
This metric measures the model’s ability to distinguish between regular and extreme events.

\subsection{Aggregation methods}

Given the previous classification problem, we must now define how to get a $[0,1]$ score from a model that outputs continuous weather variables, as a single prediction or as an ensemble of predictions.

When the model is deterministic (single prediction), from the single predicted local real-valued anomaly $\hat{x}$, a natural score $\hat{s}$ can be computed as $\hat{s} = \Phi(\hat{x})$.
Given previous considerations, $\hat{s}$ should represent the predicted quantile and we can expect $\tau = q$ to be a good threshold.

However, when the model is generative and outputs an ensemble of local anomalies $\{\hat{x}_i\}_{1 \leq i \leq n}$, the score $\hat{s}$ is not as straightforward to compute.
A simple method is to compute the mean of the local anomalies and then apply the same procedure, yet this can be suboptimal as we will discuss later.

In this study, we investigate the use of a less trivial aggregation method: the power mean.
As we are trying to predict extreme events, which are rare by definition, we might want to give more "weight" to members predicting an extreme.
The most radical way to do that would be to use the maximum over the members as the actual prediction.
However, when generating many members, this method would create too many False Positives.
We therefore used a power mean aggregation \citep{cantrellPowerMean,borweinPiAGMStudy1998} as a compromise between the average and the maximum methods.

\begin{wrapfigure}{l}{0.5\textwidth}
  \centering
  \includegraphics[width=0.9\linewidth]{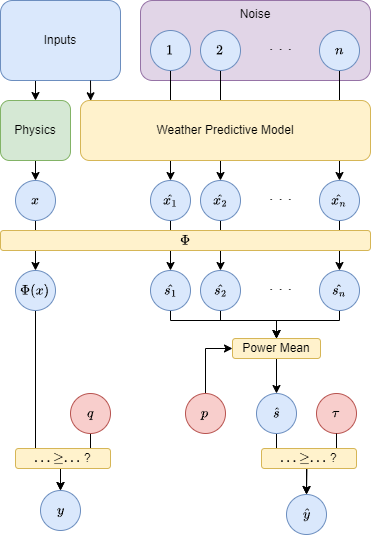}
  \caption{Aggregation method flowchart}
  \label{fig:aggregation_flow}
  \vspace{-1.5cm}
\end{wrapfigure}

Let's denote $n$ the number of members generated, $p \geq 1 $ a power exponent and $\{\hat{x}_i \}_{1 \leq i \leq n}$ the local anomalies predictions.
For each member of the ensemble, we compute a score $\hat{s}_i = \Phi(\hat{x}_i)$.
The final score $\hat{s}$ is then computed as:

\begin{equation}
  \hat{s} = \left( \frac{1}{n} \sum_{i=1}^n \hat{s}_i^p \right)^{1/p}
\end{equation}

This method is illustrated in Figure \ref{fig:aggregation_flow}.
Note that the mean is applied to member scores $\hat{s}_i$ and not directly to anomalies $\hat{x}_i$ because this operation requires positive numbers.

Previous studies already investigated the use of the power mean for ensemble prediction aggregation in various application contexts \citep{hassanNewMethodEnsemble2021} .
They demonstrated that the arithmetic mean is not always the best choice and that the optimal value of $p$ can vary depending on the application.

\subsection{Model architecture}

In order to test our aggregation method, we needed a generative model that could predict and output an ensemble of local anomalies.
We chose to use a U-Net Convolutional Neural Network-like model inspired by the work of \citet{weynImprovingDataDrivenGlobal2020} and \citet{lopez-gomezGlobalExtremeHeat2023}.
Details on the model architecture are given in appendix \ref{app:model_architecture}.
The inputs of our model are atmospheric and ground variables at days $d$, $d-1$,... $d-3$, and the outputs are surface air temperature local anomalies at days $d+1$, $d+2$, \dots, $d+12$.

To make our model generative, we followed the work of \citet{biPanguWeather3DHighResolution2022} and integrated Perlin noise \citep{perlinImageSynthesizer1985a} into the input of the deterministic baseline. 
For a given member of the ensemble, noise is randomly generated at different space scales and combined with the input through convolutions (more details in appendix \ref{app:model_architecture_generative}).
For the results presented in this paper, the model generated $n=50$ ensemble members.

\begin{wrapfigure}{r}{0.45\textwidth}
  \centering
  \vspace{-0.6cm}
  \includegraphics[width=0.95\linewidth]{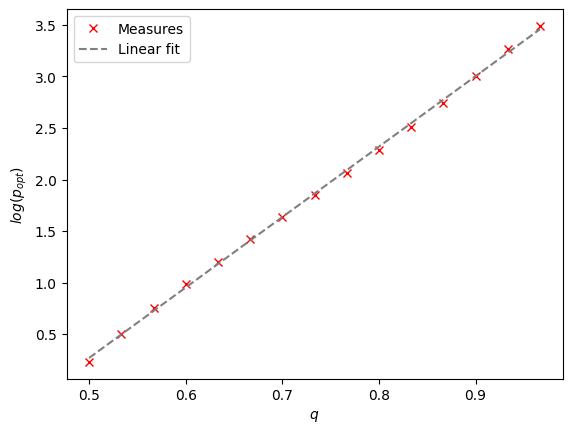}
  \vspace{-0.2cm} 
  \caption{$log(p_{opt})=f(q)$}
  \vspace{-2.5cm}
  \label{fig:linear_regression}
\end{wrapfigure}

The model was trained on data from 1990 to 2010 using a Continuous Ranked Probability Score (CRPS) \citep{berrischCRPSLearning2023} loss function.
For more details on the training procedure, see appendix \ref{app:model_architecture_training}.

\section{Results}

\subsection{Optimizing the power exponent}

To classify heat waves with a generative model, we introduced a power exponent $p$ as a hyperparameter of the aggregation method.
For a given quantile threshold $q$, we computed the AUC for different values of $p$ on the validation dataset (2010-2015).
The forecast lead time was fixed to 7 days for this experiment.
As conjectured, this function reaches a maximum at a given value of $p_{opt} \geq 1 $ and exceeds the AUC of the mean prediction method.

When computing the optimal power exponent $p_{opt}$ for different quantile thresholds, we found that this quantity increases almost perfectly exponentially with $q$ as shown in Figure \ref{fig:linear_regression}.
This exponential scaling provides a simple estimate that can be used to predict the optimal power exponent for any given quantile threshold.
Also, it emphasizes the adaptiveness of this method in its tuning when it comes to predicting extreme events, see appendix \ref{app:results_details} for more details.

\subsection{Final performance}

\begin{wrapfigure}{l}{0.5\textwidth}
  \centering
  \vspace{-0.35cm}
  \includegraphics[width=0.95\linewidth]{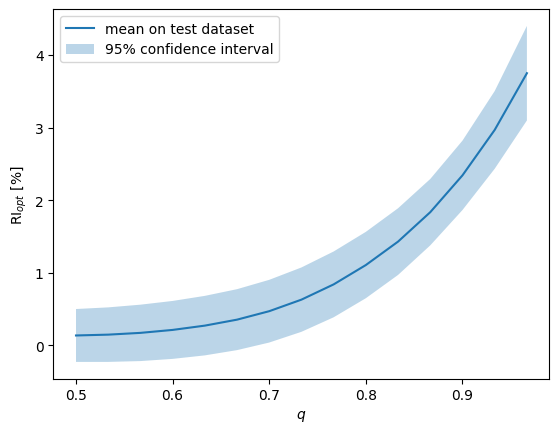}
  \caption{$\text{RI}_{opt}=f(q)$ on test dataset}
  \vspace{-0.6cm}
  \label{fig:RI_test}
\end{wrapfigure}

In the test data set (2015-2018), we computed the AUC for 7-day-ahead predictions, for different quantiles, and for both mean and power mean aggregation methods.
To measure the effectiveness of our method, we define for each quantile $q$ the relative improvement as:

\begin{equation}
  \text{RI}_{opt} = 100 \times \frac{\text{AUC}_{p_{opt}}- \text{AUC}_{\text{mean pred}}}{\text{AUC}_{\text{mean pred}}}
\end{equation}

The results presented in Figure \ref{fig:RI_test} show that the power mean aggregation method outperforms the mean prediction method for all quantiles and its effectiveness increases with the quantile threshold considered.

\begin{figure}[ht]
  \centering
  \begin{subfigure}{0.325\linewidth}
    \centering
    \includegraphics[width=\linewidth]{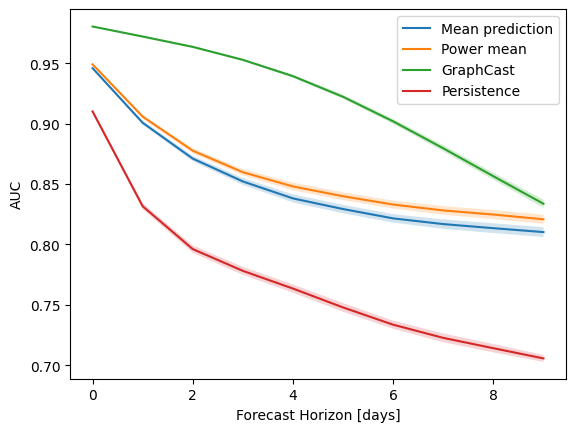}
    \caption{$q=0.8$}
  \end{subfigure}
  \begin{subfigure}{0.325\linewidth}
    \centering
    \includegraphics[width=\linewidth]{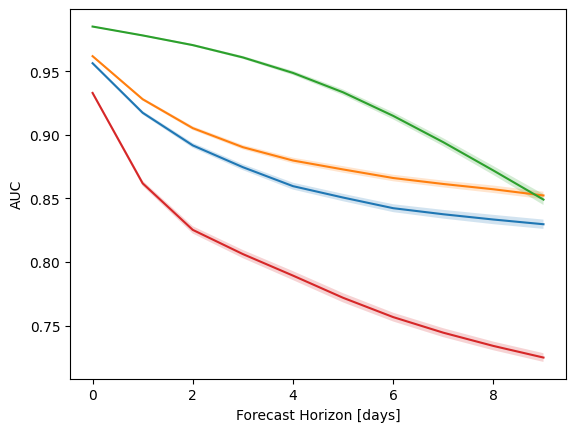}
    \caption{$q=0.9$}
  \end{subfigure}
  \begin{subfigure}{0.325\linewidth}
    \centering
    \includegraphics[width=\linewidth]{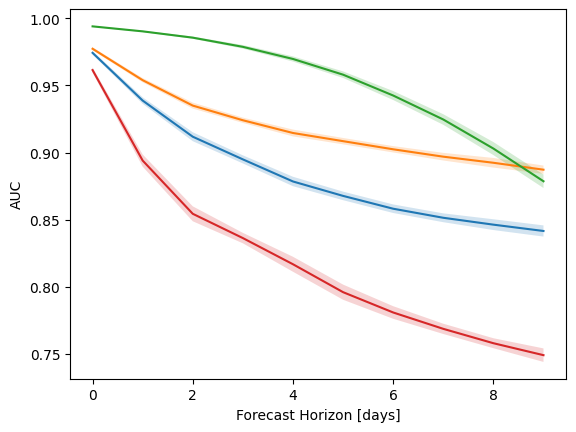}
    \caption{$q=0.98$}
  \end{subfigure}
  \caption{AUC as a function of the forecast lead time for different quantiles}
  \label{fig:test_auc}
\end{figure}

Finally, we compared the performance of our method to a state-of-the-art AI weather model: GraphCast from DeepMind (deterministic) as a reference \citep{lamGraphCastLearningSkillful2023}.
We kept the power exponents optimized on 7-day-ahead predictions but used them to compute the AUC for other forecast lead times.
We compared 4 models: the mean prediction method (ensemble predictions), the optimized power mean method (ensemble predictions), GraphCast (single prediction), and the persistence model (single prediction) as a baseline.
The results calculated for the year 2018 are presented in Figure \ref{fig:test_auc}.

Again, the results show that the power mean aggregation method outperforms the mean prediction method for all quantiles and forecast lead times, confirming the relevance of this method.
Also, they suggest that the relative improvement increases with the forecast lead time in addition to the quantile threshold.
Moreover, for high quantiles and time leads, the power mean method applied to our simple generative model even outperforms GraphCast, illustrating once again the effectiveness of this method for extreme events prediction.

\newpage

\section{Conclusion}

In this paper, we investigated the use of the power mean as an adaptive ensemble aggregation method for extreme event classification.
The results confirmed the relevance of such an approach and showed that its effectiveness increases with the level of extreme intensity considered.  

For future work, it would be interesting to apply this method to more complex and effective baseline models. 
This approach could also be generalized to any kind of weather variable to predict other extremes like droughts, heavy rains, storms…

\section*{Impact Statement} \label{impact_statement}

Extreme weather events, such as heat waves, have dramatic impacts including direct human losses, health issues, and material damages, imposing enormous costs on society. 
With climate change, these events are increasing in both frequency and intensity as stated by the Intergovernmental Panel on Climate Change (IPCC) in the 11th chapter of their 6th assessment report. \citep{intergovernmentalpanelonclimatechangeipccClimateChange20212023}

In this context, predicting extreme events is more important than ever. 
Accurate predictions can be used at various scales to trigger preparation or emergency plans, allowing society to anticipate better, be more resilient, and reduce the impact of extreme events.

The method we investigated in this study shows real promise, significantly improving the performance of our model, and easily transferable and applicable to another model. 
An increase in the accuracy of extreme event classification could help society better anticipate heat waves and other meteorological disasters, reducing their negative impact, such as mortality rates.

\section*{Reproducibility}

The data used in this study can be directly downloaded from the \href{https://weatherbench2.readthedocs.io/en/latest/data-guide.html}{WeatherBench2 website}.\citep{raspWeatherBenchBenchmarkNext2024} 

The appendix details the data used and the model architecture implemented for reproducibility purposes.

Experiments were conducted on a single GPU of 16GB.
Storing the data and the models requires about 20 GB of disk space.
Training the model takes a few hours on a single GPU.
Testing a model to reproduce the results presented in this paper takes less than an hour on a single GPU.

\section*{Limitations}

This study has several limitations that should be considered:
\begin{itemize}
  \item \textbf{Simplified definition of extreme events:} The definition of extreme events used in this study is quite simple and does not capture all the complexity of real extreme events.
  The work of \citet{zscheischlerTypologyCompoundWeather2020} illustrates this complexity and the need for more advanced definitions.
  \item \textbf{Univariate problems:} Many extreme events are multivariate and depend on the interaction of several variables.
  Working with such problems requires more sophisticated climatology models, like copula-based distributions.
  The work presented in this paper only focuses on a single variable: surface air temperature.
  \item \textbf{Static climatology:} The anomalies considered in this study to define extremes and train models are based on a static climatology that does not take into account climate change.
  This is a major limitation as climate change is expected to significantly impact extreme events.
  For future work, it would be interesting to consider a dynamic climatology that evolves with time.
  \item \textbf{Metric choice:} The AUC metric used in this study is a general metric for binary classification problems but is not necessarily the best choice.
  To measure the performance of extreme events predictors from a more application-based perspective, socio-economic considerations could be integrated.
\end{itemize}

\newpage

\bibliographystyle{unsrtnat}
\bibliography{used_refs, softwares}

\appendix

\section{Data and grid system} \label{app:cube_sphere_grid}

The data we downloaded for this study included solar irradiance, atmospheric variables such as geopotential and wind components at different pressure levels, and ground variables such as surface air temperature and soil water content.
This choice of data to predict surface air temperature was inspired by the work of \citet{lopez-gomezGlobalExtremeHeat2023}.
The full list of variables and the levels at which they were sampled are presented in Table \ref{tab:variables}.

\begin{table}[ht]
  \centering
  \begin{tabular}{lll}
    \toprule
    \textbf{Variable} & \textbf{Units} & \textbf{Levels} \\
    \midrule
    Surface air temperature & $K$ & 2 m above surface \\
    Geopotential & $m$ & 500, 700 hPa \\
    Wind zonal component  & $m \cdot s^{-1}$ & 500, 700 hPa \\
    Wind meridional component  & $m \cdot s^{-1}$ & 500, 700 hPa \\
    Wind speed  & $m \cdot s^{-1}$ & 500, 700 hPa \\
    Wind vorticity & $s^{-1}$ & 500, 700 hPa \\
    Specific humidity & $kg \cdot kg^{-1}$ & 500, 700 hPa \\
    Incoming solar radiation & $W \cdot m^{-2}$ & top of the atmosphere \\
    Soil water content & $m^3 \cdot m^{-3}$ & 0-7 cm depth\\
    \bottomrule
  \end{tabular}
  \vspace{0.2cm}
  \caption{Variables}
  \vspace{-0.2cm}
  \label{tab:variables}
\end{table}

Additionally, we also used three static fields:
\begin{itemize}
  \vspace{-0.2cm}
  \setlength\itemsep{0em}
  \item Latitude: $[-90,90]$ degrees
  \item Longitude: $[0, 360]$ degrees
  \item Land-sea mask: $[0,1]$ scalar field indicating the proportion of land in each grid cell.
\end{itemize}

The data were downloaded from the WeatherBench2 dataset \citep{raspWeatherBenchBenchmarkNext2024} and are originally from the ERA5 reanalysis data from the European Centre for Medium-Range Weather Forecasts (ECMWF) \citep{hersbachERA5GlobalReanalysis2020}.

\begin{wrapfigure}{r}{0.4\textwidth}
  \centering
  \vspace{0cm}
  \includegraphics[width=0.9\linewidth]{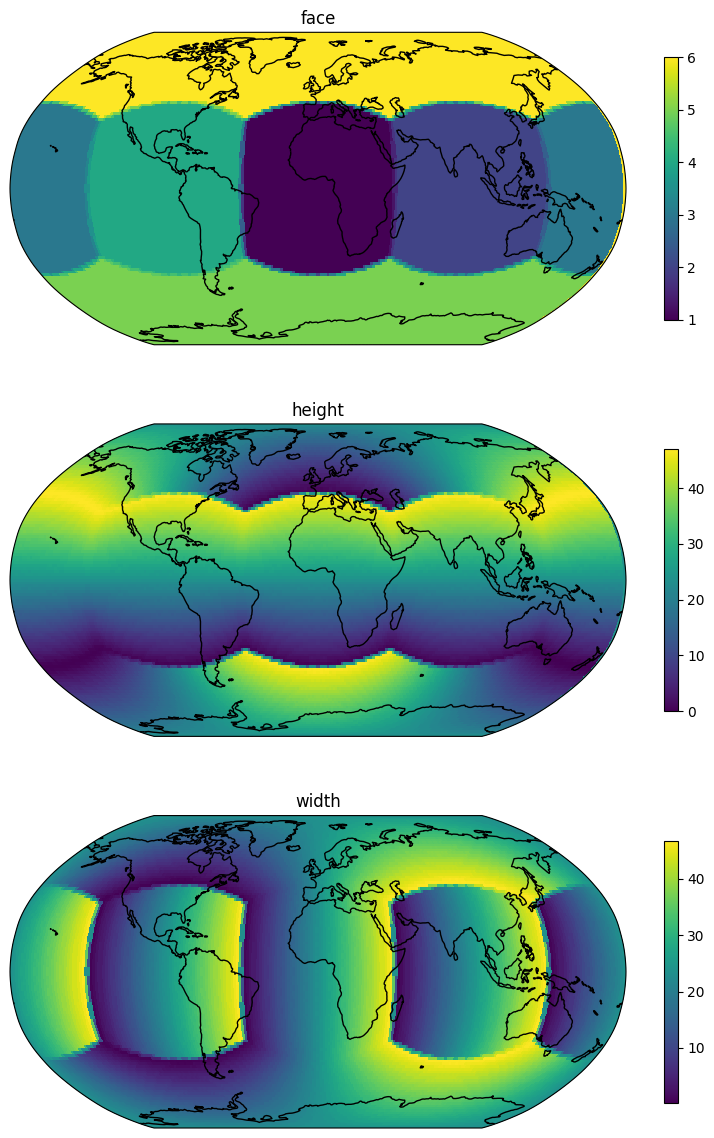}
  \caption{Cube Sphere Grid}
  \vspace{-1cm}
  \label{fig:cube_sphere}
\end{wrapfigure}

Following the work of \citet{weynImprovingDataDrivenGlobal2020}, we regridded the data to a 6x48x48 gnomonic equiangular cubic sphere grid.
This grid system has a few advantages:
\begin{enumerate}
  \item It avoids pole singularities, which are problematic for many grid systems.
  \item It gives a more uniform representation of the Earth's surface.
  With the latitude-longitude grid, data for the poles is over-represented, this is not the case with the cube sphere grid.

  \item It allows the use of a special case of Convolutional Neural Networks (CNNs) for spatial data.
\end{enumerate}

The cube sphere grid is based on a 3D cube where each face is a square of 48x48 pixels.
In this grid system, spatial coordinates are the face number $f \in \{1,...,6\}$ and the pixel heights $h \in \{1,...,48\}$ and width $w \in \{1,...,48\}$ on the face.
The grid is illustrated in Figure \ref{fig:cube_sphere}.

To regrid the data from the latitude-longitude grid to the cubic sphere grid, we used the ``TempestRemap'' software \citep{ullrichArbitraryOrderConservativeConsistent2015,ullrichArbitraryOrderConservativeConsistent2016}.
It concretely generates a weight matrix that allows us to interpolate the data from one grid to the other.
We also used k-nearest neighbors interpolation with $k=4$ for faster and simpler regridding when necessary.

\newpage

\section{Model architecture} \label{app:model_architecture}

To experiment with the power mean aggregation method, we developed a simple generative model that could output an ensemble of local anomalies.
Our implementation mainly follows the work of \citet{weynImprovingDataDrivenGlobal2020} and \citet{lopez-gomezGlobalExtremeHeat2023}.
Here is an overview of the model architecture, with more details in the following sections.

\begin{figure}[ht]
  \centering
  \includegraphics[width=\linewidth]{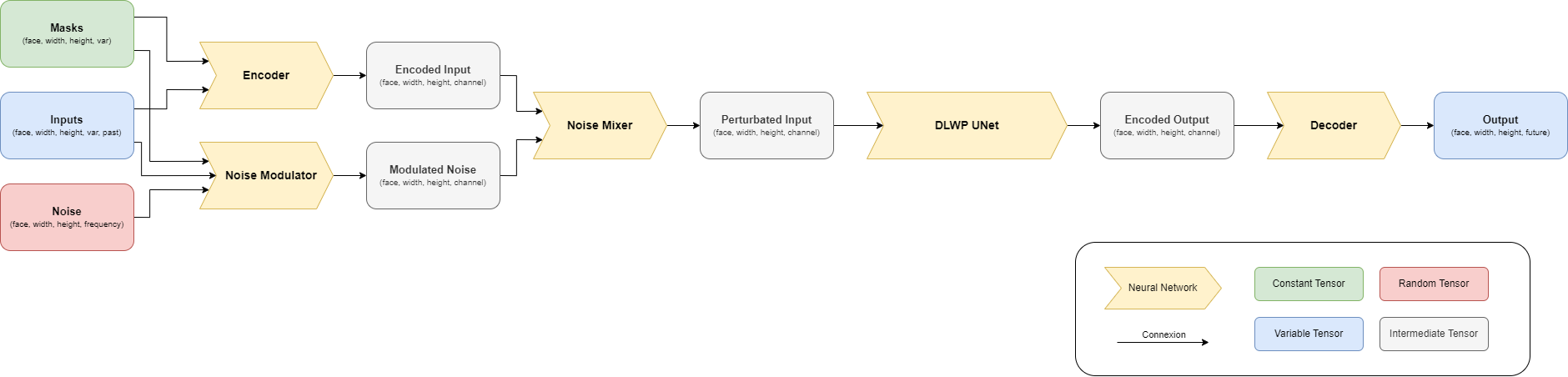}
  \caption{Model architecture overview}
  \label{fig:architecture}
\end{figure}

\subsection{Deterministic baseline}  \label{app:model_architecture_baseline}

For the needs of our study, we needed a relatively light ML model that could output local anomalies of surface air temperature at different forecast lead times (1 to 12 days).
We chose to use a Convolutional Neural Network like model inspired by the work of \citet{weynImprovingDataDrivenGlobal2020} and modified by \citet{lopez-gomezGlobalExtremeHeat2023}.
Their models are based on the U-Net architecture \citep{ronnebergerUNetConvolutionalNetworks2015} and adapted to the cube sphere grid (see appendix \ref{app:cube_sphere_grid}).

Their main idea is to apply convolutions on each face of the cube sphere grid using shared kernel weights for equatorial and polar faces.
This way, the model can learn different physics for poles and equator faces while maintaining a reasonable number of parameters.

In the U-Net part of the model, as the data goes down on resolution, the convolutions represent larger scales interactions and transport.
The skipped connections, characteristic of the U-Net architecture, allow the model to maintain a high resolution representation of the data while still learning the large scale features.

The inputs of our models are atmospheric and ground variables at day $d$, $d-1$, ... $d-3$. 
They are scaled using the global climatology mean and standard deviation.
This global scaling ensure that gradients are not distorted as they play a crucial role in the physics.
As we want to predict the surface air temperature local anomaly, this variable is also given as input to the model.
Given that the physics depends on the latitude, longitude and is significantly different on continents and oceans, we also give these three spatial masks as input to the model.
Finally, the model outputs surface air temperature local anomalies at day $d+1$, $d+2$, $...$ , $d+12$.

\subsection{Generative adaptation} \label{app:model_architecture_generative}

To investigate ensemble aggregation methods, we needed a generative model that could output an ensemble of local anomalies.
For that, we followed the work of \citet{biPanguWeather3DHighResolution2022} and added Perlin noise to the input of the deterministic baseline.
We adapted the generation of this kind of noise to the cube sphere and modified it to better capture extremes.
Also, we did not simply add the noise to the input but modulated it using the model itself.

\begin{figure}[ht]
  \centering
  \includegraphics[width=0.75\linewidth]{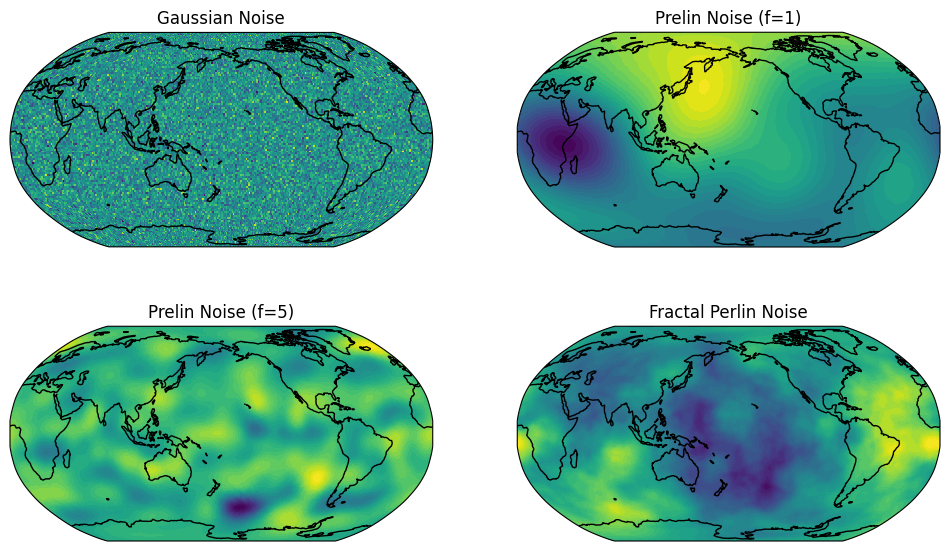}
  \caption{Gaussian Noise, Perlin Noises for different frequencies and Fractal Perlin Noise}
  \label{fig:noises}
\end{figure}

Perlin noise is a type of gradient noise developed by Ken Perlin and introduced in 1985 in a paper called ``An Image Synthesizer'' \citep{perlinImageSynthesizer1985a}.
On the contrary to white noise (pixel-wise Gaussian Noise), As shown in Figure \ref{fig:noises}, Perlin noise is smooth, continuous and has a spatial coherence that makes it suitable for generating natural-looking textures, or in our use case, weather variable fields.

\begin{figure}[ht]
  \centering
  \includegraphics[width=0.8\linewidth]{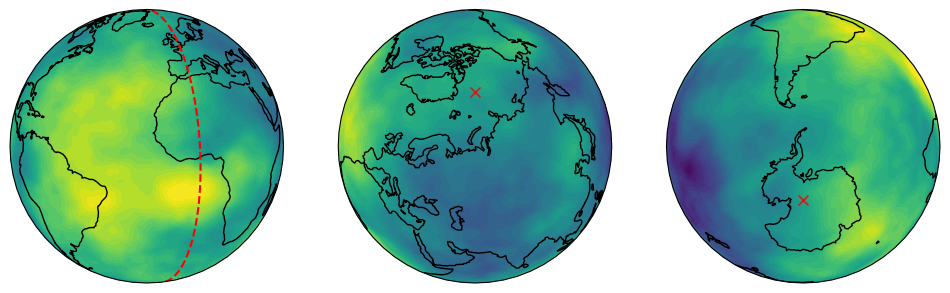}
  \caption{Generated Fractal Perlin Noise and its continuity on the globe}
  \label{fig:noise_continuity}
\end{figure}

To ensure continuity across the globe, we generated 3D Perlin Noise on the $[0,1]^3$ cube and selected the 2D slice corresponding to the Earth surface.
An example of Perlin Noise generated on the globe is shown in Figure \ref{fig:noise_continuity}.

The general approach to generating Perlin noise is detailed in Ken Perlin's 1999 talk \citep{perlinMAKINGNOISE1999} and illustrated on this website: \url{https://en.wikipedia.org/wiki/Perlin_noise#Algorithm_detail}.

The algorithm we concretely used was adapted from the work of \citet{vigierPvigierPerlinnumpy2024}.
Our contribution was to randomize the gradient vectors amplitude with a log-normal distribution to better capture extremes.
Indeed, default Perlin Noise is bounded between -1 and 1, but for our use case, we wanted a noise that could go beyond these bounds.

To represent different scales of randomness, we generated Perlin noise at different frequencies and combined them to create a sort of fractal noise using a convolutional layer as an amplitude modulator.
This noise modulator layer uses the same input as the deterministic model and learns how to generate a spatially coherent noise adapted to the data.
Finally the noise is mixed into the input using a simple 1x1 convolutional layer as illustrated in Figure \ref{fig:architecture}.

\subsection{Model implementation}

Our model is implemented in "Flax" \citep{flax2020github} as a ML high-level API based on "JAX" \citep{jax2018github}, a numerical computing library.
Following the work of \citet{lopez-gomezGlobalExtremeHeat2023}, the different types of layers used in the model are:
\begin{itemize}
  \item \textbf{Conv}: 2D dilated convolutions layer \citep{yuMultiScaleContextAggregation2016} with a kernel size of 3x3 or 1x1, a stride of 1 and a specified dilation factor. 
  Dilation is used to increase the receptive field of the model without increasing the number of parameters. 
  Depending on the kernel size and dilation factor, padding is added to maintain the same spatial resolution with data from other faces of the cube sphere grid. 
  The notation Conv(32, 3x3, 1) means a convolutional layer with 32 filters, a kernel size of 3x3, and a dilation of 1.
  \item \textbf{PreLU}: Parametric ReLU activation functions with shared parameters along all dimensions except the channel axis.
  \item \textbf{MaxPooling}: 2x2 max pooling layers.
  \item \textbf{Reshaping}: Layer reshaping a tensor to a given shape.
  \item \textbf{Concat}: Concatenation of two tensors along the channel axis.
  \item \textbf{Product}: Element-wise product of two tensors.
  \item \textbf{Sum}: Sum of a tensor along a given axis.
\end{itemize}

The different blocks of our model, introduced in Figure \ref{fig:architecture}, are made of the following layers:
\setlength{\tabcolsep}{1mm} 
\begin{table}[ht]
  \centering
  \begin{minipage}{0.45\textwidth}
    \centering

    \begin{tabular}{lc}
      \toprule
      \textbf{Layer}  & \textbf{Output shape}\\
      \midrule
      Input & (6,48,48,27) \\
      \midrule
      Conv(32,3x3,1) & (6,48,48,32)\\
      PreLU  & (6,48,48,32)\\
      Conv(32,3x3,1) &  (6,48,48,32) \\
      PreLU &  (6,48,48,32)\\
      \bottomrule
      Output &  (6,48,48,32) \\
    \end{tabular}
    \vspace{0.2cm}
    \caption{Encoder}

    \vspace{0.2cm}
    \begin{tabular}{lcc}
      \toprule
      \textbf{Layer} & \textbf{Output shape} & \textbf{Ref}\\
      \midrule
      Input & (6,48,48,27) & \\
      \midrule
      Conv(64,3x3,1) & (6,48,48,64)&\\
      PreLU &  (6,48,48,64) & \\
      Reshaping & (6,48,48,16,4) & [1]\\
      Perlin Noise & (6,48,48,16,4) & [2]\\
      Product [1] x [2] &  (6,48,48,16,4) & \\
      Sum [last axis] & (6,48,48,16) & \\
      \bottomrule
      Output & (6,48,48,16) & \\
    \end{tabular}
    \vspace{0.2cm}
    \caption{Noise Modulator}

    \vspace{0.2cm}
    \begin{tabular}{lc}
      \toprule
      \textbf{Layer} & \textbf{Output shape}\\
      \midrule
      Input A & (6,48,48,32) \\
      Input B & (6,48,48,16) \\
      \midrule
      Concat [A] \& [B] &(6,48,48,48) \\
      Conv(32,1x1,1) & (6,48,48,32) \\
      PreLU & (6,48,48,32)\\
      \bottomrule
      Output & (6,48,48,32) \\
    \end{tabular}
    \vspace{0.2cm}
    \caption{Noise Mixer}

    \vspace{0.2cm}
    \begin{tabular}{lc}
      \toprule
      \textbf{Layer} & \textbf{Output shape}\\
      \midrule
      Input & (6,48,48,32) \\
      \midrule
      Conv(12,1x1,1) & (6,48,48,12)\\
      \bottomrule
      Output & (6,48,48,12) \\
    \end{tabular}
    \vspace{0.2cm}
    \caption{Decoder}

  \end{minipage}
  \hspace{0.05\textwidth} 
  \begin{minipage}{0.45\textwidth}
    \centering
    \begin{tabular}{lcc}
      \toprule
      \textbf{Layer} & \textbf{Output shape} & \textbf{Ref}\\
      \midrule
      Input &  (6,48,48,32) & [1]\\
      \midrule
      MaxPooling & (6,24,24,64) & \\
      Conv(64,3x3,1) & (6,24,24,64) & \\
      PreLU  & (6,24,24,64) & \\
      Conv(64,3x3,2) & (6,24,24,64) & \\
      PreLU & (6,24,24,64) & [2]\\
      \midrule
      MaxPooling &(6,12,12,64) & \\
      Conv(128,3x3,1) & (6,12,12,128) & \\
      PreLU & (6,12,12,128) & \\
      Conv(128,3x3,4) & (6,12,12,128) & \\
      PreLU & (6,12,12,128) & \\
      Conv(128,3x3,8)& (6,12,12,53) & \\
      PreLU & (6,12,12,64) & [3]\\
      \midrule
      Up-Sampling [3] & (6,24,24,64) & [4]\\
      Concat [2] \& [4] & (6,24,24,128) & \\
      Conv(64,3x3,1) & (6,24,24,64) & \\
      PreLU & (6,24,24,64) & \\
      Conv(64,3x3,2) & (6,24,24,64) & \\
      PreLU & (6,24,24,64) & \\
      Conv(32,3x3,4) & (6,24,24,32) & \\
      PreLU & (6,24,24,32) & [5]\\
      \midrule
      Up-Sampling [4] & (6,48,48,32) & [6]\\
      Concat [1] \& [6] & (6,48,48,64) & \\
      Conv(32,3x3,1) & (6,48,48,32) & \\
      PreLU & (6,48,48,32) & \\
      Conv(32,3x3,2) & (6,48,48,32) & \\
      PreLU & (6,48,48,32) & \\
      Conv(32,3x3,1) & (6,48,48,32) & \\
      PreLU & (6,48,48,32) & \\
      \bottomrule
      Output & (6,48,48,32) & \\
    \end{tabular}
    \vspace{0.2cm}
    \caption{DLWP U-Net}
  \end{minipage}
\end{table}

Our model contains approximately 1.2 million learnable parameters. In comparison, GraphCast \citep{lamGraphCastLearningSkillful2023} has around 36.7 million parameters.

\subsection{Training procedure}  \label{app:model_architecture_training}

To train our model to predict coherent ensembles of local anomalies, we used a probabilistic loss function: the Continuous Ranked Probability Score (CRPS) loss function.
The CRPS is a proper scoring rule for probabilistic forecasts defined as:
\begin{equation}
  \text{CRPS} = \int_{-\infty}^{+\infty} (\text{P}(\hat{x} \leq x) - \mathds{1}_{x \leq x_i})^2 dx
\end{equation}
where $\hat{x}$ is the predicted local anomaly seen as a random variable, $x$ is the true local anomaly and $\mathds{1}$ is the indicator function.
Numerically, we approximated the loss as:
\begin{equation}
  \mathcal{L} = \frac{1}{N} \sum_{i=1}^N \left[ \frac{1}{n} \sum_{j=1}^n |\hat{x}_{i,j} - x_i| - \frac{1}{2n^2} \sum_{j=1}^n \sum_{k=1} |\hat{x}_{i,j} - \hat{x}_{i,k}|\right]
\end{equation}
where $1 \leq i \leq N$ represents a given time, location and forecast lead time and $1 \leq j \leq n$ the member index in the ensemble ($n$ is the ensemble size). $x_i$ is the true local anomaly and
$\hat{x}_{i,j}$ are the predicted anomalies.
This approximation comes from the fact that the CRPS can be expressed as:
\begin{equation}
  \text{CRPS} = \mathbb{E} \left[|\hat{X}-x|\right] - \frac{1}{2} \mathbb{E} \left[|\hat{X}-\hat{X'}|\right]
\end{equation}
where $\hat{X}$ and $\hat{X'}$ are two independent random variables with the same distribution as $\hat{x}$,
according to the work of \citet{gneitingStrictlyProperScoring2007}.

The model was trained using the Adam optimizer with a learning rate of $10^{-3}$ until the loss on the validation dataset stopped decreasing.
The training was done on a single GPU with 16GB of memory and took a few hours.

\newpage

\section{Results details} \label{app:results_details}

\subsection{Overall predictive performances}

A common way to evaluate the performance of a weather predictive model is to compute the Root Mean Squared Error (RMSE) between the predicted and true temperatures.
For a given forecast lead time, we computed the RMSE across the whole test dataset as:
\begin{equation}
  \text{RMSE} = \sqrt{\frac{1}{N} \sum_{i=1}^N (\hat{x}_i - x_i)^2}
\end{equation}
where $N$ is the number of members ($\simeq 365 \times 6 \times 48 \times 48 \simeq 5 \cdot 10^6 $), $\hat{x}_i$ are the predicted global anomalies and $x_i$ the true global anomalies.
The lower the RMSE, the better the model.

\begin{figure}[h]
  \centering
  \includegraphics[width=0.6\linewidth]{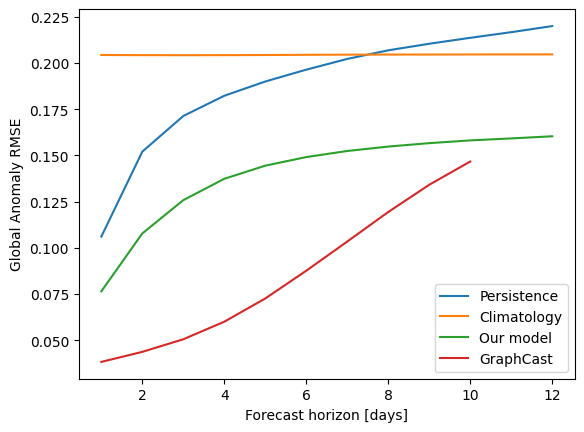}
  \caption{Global anomaly RMSE on test dataset}
  \label{fig:rmse}
\end{figure}

We computed the RMSE of temperature global anomalies for different forecast lead times and different models:
\begin{itemize}
  \item Persistence: a simple model that predicts the temperature at time $d+k$ to be the same as at time $d$.
  \item Climatology: always predicts the local mean temperature at any time of the year and location.
  \item Our model: the generative model we introduced that outputs an ensemble of local anomalies. The mean prediction is considered for this RMSE computation.
  \item GraphCast: GraphCast model \citep{lamGraphCastLearningSkillful2023} that outputs a single prediction.
\end{itemize}
Results are presented in Figure \ref{fig:rmse}.

As expected, our generative model performs better than the persistence baseline and remained predictive even up to 12 days ahead, as its RMSE stays below the climatology reference. 
Also, and not surprisingly, its RMSE was always higher than GraphCast's, a more complex model well-suited for short-term weather forecasting.

\subsection{Optimal power exponent}

To find the optimal power exponent $p_{opt}$ for a given quantile threshold $q$, we computed the AUC for different values of $p$ on the validation dataset (2010-2015).
As mentioned in the main text, we fixed the forecast lead time to 7 days for simplicity. 

\begin{figure}
  \centering
  \includegraphics[width=0.7\linewidth]{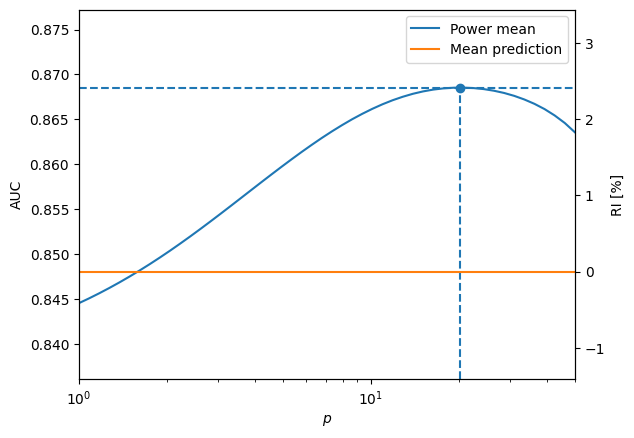}
  \caption{$\text{AUC} = f(p)$ for $q=0.9$}
  \label{fig:auc_vs_p}
\end{figure}

We defined the optimal power exponent $p_{opt}$ as the value of $p$ that maximizes the AUC and the relative improvement as:
\begin{equation*}
  \text{RI} = 100 \times \frac{\text{AUC}_{p}- \text{AUC}_{\text{mean pred}}}{\text{AUC}_{\text{mean pred}}}
\end{equation*}

For the curve presented in Figure \ref{fig:auc_vs_p} ($q=0.9$), the optimal power exponent is $p_{opt} \simeq  18.3$ and the relative improvement is $\text{RI}_{opt} \simeq 2.67\%$.
We can also see that the AUC for $p=1$ is not the same as the AUC for the mean prediction method, because of the non-linearity of the normal distribution quantile function $\Phi$.

Repeating this process for different quantiles, we obtained the following results:

\begin{figure}[ht]
  \centering
  \includegraphics[width=0.8\linewidth]{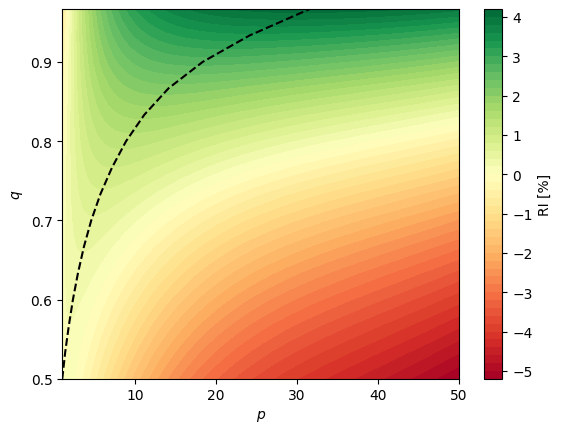}
  \caption{$\text{RI}=f(p,q)$}
  \label{fig:2D_plot}
\end{figure}

From this plot, we can see that the optimal power exponent $p_{opt}$ seems to be exponentially increasing with the quantile threshold $q$.
This assumption is confirmed by the linear regression presented in the main section in Figure \ref{fig:linear_regression}.
It also suggests that the relative improvement $\text{RI}$ increases with the quantile threshold $q$, the result also confirmed on the test dataset as shown in Figure \ref{fig:RI_test}.

\end{document}